\documentclass[letterpaper, 10 pt, conference]{ieeeconf}

\IEEEoverridecommandlockouts                              % This command is only needed if
\usepackage{graphicx} % for pdf, bitmapped graphics files
\usepackage{amsmath} % assumes amsmath package installed
\usepackage{amssymb}  % assumes amsmath package installed
\usepackage{bm}
\usepackage{capt-of}
\usepackage[ruled]{algorithm2e}
\usepackage{booktabs}
\usepackage{multirow}
\usepackage{textcomp}
\usepackage[table]{xcolor}% http://ctan.org/pkg/xcolor
\usepackage{array,colortbl}

\newcolumntype{P}[1]{>{\centering\arraybackslash}p{#1}}

\usepackage{stfloats}
\usepackage{cuted}
\usepackage[bookmarks=false,pdfborder={0 0 0}]{hyperref}
\hypersetup{hidelinks}

\title{\LARGE \bf
\vspace{10pt}
FeasibleCap: Real-Time Embodiment Constraint Guidance\\for In-the-Wild Robot Demonstration Collection
\vspace{-10pt}
}

\author{Zi Yin, Fanhong Li, Yun Gui, Jia Liu$^{*}$\\[2pt]
{\normalsize Tsinghua University, Beijing, China}\\[1pt]
{\small $^{*}$Corresponding author}
}

\begin{document}

\twocolumn[{%
    \renewcommand\twocolumn[1][]{#1}%
    \maketitle
    \vspace{-4mm}
    \begin{center}
        \includegraphics[width=\textwidth]{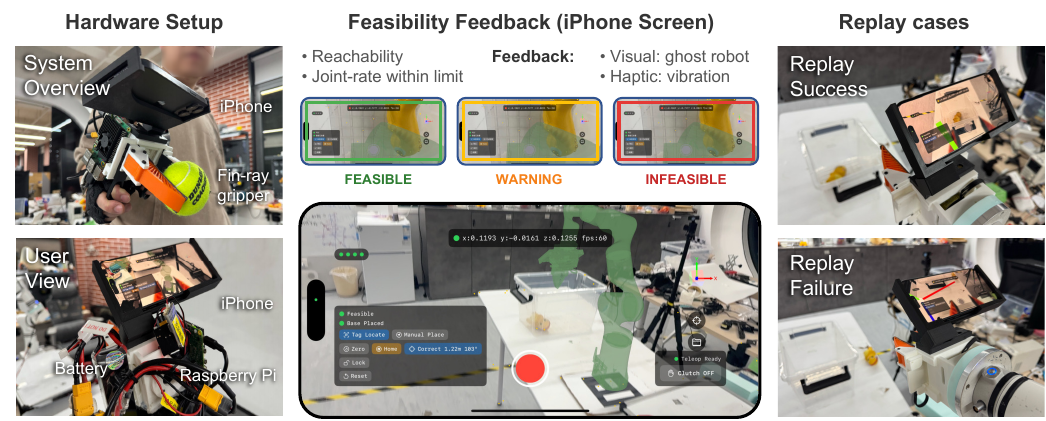}
        \captionof{figure}{FeasibleCap overview. An iPhone is mounted on a handheld gripper, providing real-time feasibility feedback via an on-screen indicator. The indicator turns red as the end-effector approaches workspace boundaries or joint-rate limits, guiding demonstrators to stay within the target robot's executable region.}
        \label{fig:teaser}
    \end{center}
    \vspace{-2mm}
}]

\thispagestyle{empty}
\pagestyle{empty}

%%%%%%%%%%%%%%%%%%%%%%%%%%%%%%%%%%%%%%%%%%%%%%%%%%%%%%%%%%%%%%%%%%%%%%%%%%%%%%%%
\begin{abstract}
Gripper-in-hand data collection decouples demonstration acquisition from robot hardware, but whether a trajectory is executable on the target robot remains unknown until a separate replay-and-validate stage. Failed demonstrations therefore inflate the effective cost per usable trajectory through repeated collection, diagnosis, and validation. Existing collection-time feedback systems mitigate this issue but rely on head-worn AR/VR displays, robot-in-the-loop hardware, or learned dynamics models; real-time executability feedback has not yet been integrated into the gripper-in-hand data collection paradigm. We present \textbf{FeasibleCap}, a gripper-in-hand data collection system that brings real-time executability guidance into robot-free capture. At each frame, FeasibleCap checks reachability, joint-rate limits, and collisions against a target robot model and closes the loop through on-device visual overlays and haptic cues, allowing demonstrators to correct motions during collection without learned models, headsets, or robot hardware. On pick-and-place and tossing tasks, FeasibleCap improves replay success and reduces the fraction of infeasible frames, with the largest gains on tossing. Simulation experiments further indicate that enforcing executability constraints during collection does not sacrifice cross-embodiment transfer across robot platforms. Hardware designs and software are available at \url{https://github.com/aod321/FeasibleCap}.
\end{abstract}

%%%%%%%%%%%%%%%%%%%%%%%%%%%%%%%%%%%%%%%%%%%%%%%%%%%%%%%%%%%%%%%%%%%%%%%%%%%%%%%%
\section{Introduction}

Gripper-in-hand data collection has made it practical to acquire large-scale demonstration datasets without requiring robot hardware during capture, enabling researchers to scale data collection across diverse environments~\cite{chi2024umi}. However, removing the robot from the collection loop does not make the overall process cheap. In this paradigm, whether a demonstrator's motion is actually executable by the target robot remains unknown until a separate replay-and-validate stage. Failed demonstrations incur the full cost of collection, replay, diagnosis, and re-collection, raising the effective cost per usable trajectory well beyond what the collection effort alone would suggest. This problem becomes especially severe as tasks get faster or more dynamically demanding.

The appeal of gripper-in-hand collection rests on two properties. First, collection scales independently without occupying robot resources. Second, the gripper itself is the end-effector, so no retargeting from human motion is required and the physical correspondence between demonstration and execution is preserved. UMI~\cite{chi2024umi} and a growing family of variants extend this paradigm with richer sensing, tactile feedback, and multi-view capture~\cite{wu2024fastumi,xu2025dexumi,cheng2026tacumi,rayyan2025mvumi,activeumi2025}. Yet because no robot is present during collection, demonstrators have no awareness of the target robot's kinematic constraints. Workspace violations, joint-rate exceedances, and collisions are all invisible at collection time and only surface during replay. This is particularly consequential for fast actions such as tossing, where joint-rate limits are sensitive to small speed differences and replay failures are common, yet these boundary-case motions are precisely the ones that matter most for policy robustness.

Prior work has established that collection-time feasibility feedback is effective. ARCap~\cite{chen2024arcap} overlays a virtual robot in a head-mounted display and issues visual and haptic warnings as joint or speed limits are approached, substantially improving replay success. ARMADA~\cite{armada2024} and ARMimic~\cite{armimic2025} leverage Apple Vision Pro to visualize a virtual robot during collection. FABCO~\cite{fabco2025} computes real-time feasibility scores from pre-trained dynamics models and incorporates them into feasibility-weighted behavior cloning. Collectively, these systems show that guiding demonstrators during capture improves data quality. However, they rely on head-worn AR/VR devices, robot-in-the-loop hardware, or learned dynamics models trained from robot data, and therefore cannot be directly applied to the gripper-in-hand paradigm. Despite the rapid adoption of gripper-in-hand collection, real-time executability feedback has not yet been integrated into this paradigm.

We present \textbf{FeasibleCap}, a gripper-in-hand data collection system that brings real-time executability guidance into robot-free capture. An iPhone is mounted on the gripper with its camera facing outward and its screen facing the demonstrator. At each frame, the system estimates the end-effector pose via ARKit, solves inverse kinematics on-device against a target robot model, checks reachability, joint-rate limits, and collisions, and delivers immediate feedback through an AR ``ghost arm'' rendered on screen and haptic vibration. Demonstrators can correct motions on the fly rather than discovering failures only at replay time. FeasibleCap requires no learned dynamics model, no head-worn display, and no robot hardware during collection. To our knowledge, it is the first system to provide collection-time executability feedback within the gripper-in-hand paradigm.

Our contributions are threefold:
\begin{itemize}
    \item We identify the executability gap in robot-free gripper-in-hand demonstration pipelines: collected trajectories cannot be validated until a costly replay stage, yet no existing feedback mechanism is compatible with this paradigm.
    \item We present FeasibleCap, which brings collection-time feasibility guidance into gripper-in-hand capture without head-mounted displays, robot hardware, or learned dynamics models.
    \item We show that such guidance substantially improves replay success---with the largest gains on dynamically demanding tasks---while preserving cross-embodiment transferability.
\end{itemize}
%%%%%%%%%%%%%%%%%%%%%%%%%%%%%%%%%%%%%%%%%%%%%%%%%%%%%%%%%%%%%%%%%%%%%%%%%%%%%%%%
\section{Related Work}

\subsection{Handheld and Robot-Free Demonstration Collection}

Handheld gripper interfaces have emerged as a scalable alternative to teleoperation by decoupling data collection from robot hardware. UMI~\cite{chi2024umi} establishes the gripper-in-hand paradigm, combining SLAM-based pose tracking with post-hoc kinematic filtering to discard infeasible demonstrations. A family of variants addresses specific limitations: Fast-UMI~\cite{wu2024fastumi} replaces SLAM with onboard VIO, DexUMI~\cite{xu2025dexumi} extends the concept to hand exoskeletons, TacUMI~\cite{cheng2026tacumi} integrates visuotactile sensing, MV-UMI~\cite{rayyan2025mvumi} adds multi-view capture, and ActiveUMI~\cite{activeumi2025} augments collection with head-mounted active perception. LEGATO~\cite{seo2024legato} generalizes the approach to cross-embodiment transfer across Franka, Spot, and quadruped platforms via a motion-invariant representation. Across all these systems, executability is assessed only after collection; demonstrators receive no guidance during capture, so the cost of replay failures and re-collection remains unavoidable at the source. RAPID~\cite{yin2026rapid} is a lightweight and compact handheld collection platform that supports rapid reconfiguration of gripper types and sensor modalities, enabling low-cost iteration across task setups. Its direct in-hand form factor---where the same gripper used during collection mounts directly onto the robot arm for replay---further closes the embodiment gap and makes it well suited for integrating additional sensing and feedback capabilities. Like the systems above, however, RAPID does not provide executability feedback during collection, and demonstration quality still depends on post-hoc replay validation.

\subsection{Collection-Time Feedback for Demonstration Quality}

Prior work has demonstrated that collection-time feedback can substantially improve the executability of collected demonstrations. ARCap~\cite{chen2024arcap} overlays a virtual robot in a VR headset and triggers visual warnings and haptic vibration when joint or speed limits are approached, increasing replay success by over 40\% in user studies. ARMADA~\cite{armada2024} and ARMimic~\cite{armimic2025} use Apple Vision Pro to visualize a virtual robot during collection, with ARMADA reporting replay success rates of 71.1\% with feedback versus 1.3\% without. FABCO~\cite{fabco2025} computes real-time feasibility scores from pre-trained forward and inverse dynamics models, provides color-coded visual feedback and haptic blocking, and incorporates these scores into feasibility-weighted behavior cloning for downstream training. JoyLo~\cite{joylo2025} achieves high replay success through joint-to-joint teleoperation with impedance feedback, effectively bringing robot hardware into the collection loop to guarantee executability. These works collectively establish that collection-time guidance improves data quality, yet all require head-worn AR/VR displays, robot-in-the-loop hardware, or learned dynamics models trained from robot execution data. As a result, real-time executability feedback has not been integrated into this lightweight, retargeting-free, physically grounded paradigm.

Table~\ref{table:comparison} positions FeasibleCap among recent collection-time feedback systems. ARCap, ARMADA, and ARMimic all require head-worn displays (VR headsets or Apple Vision Pro) to visualize the virtual robot, adding cost, setup complexity, and ergonomic burden to the collection process. JoyLo achieves high replay success through joint-to-joint impedance teleoperation but requires the physical robot to be active during collection, forfeiting the scalability benefit of robot-free capture. FABCO also uses a hand-mounted demonstration interface and operates without a headset, but its feasibility estimation relies on learned forward and inverse dynamics models trained from robot execution data, whereas FeasibleCap computes feasibility analytically from the target robot's URDF without any learned model. FeasibleCap is, to our knowledge, the only system that simultaneously eliminates the need for a head-worn display, robot-in-the-loop hardware, and learned dynamics models, while still delivering real-time feasibility feedback during collection.

\begin{table}[t]
    \small
    \caption{\footnotesize{\textbf{Comparison of collection-time feedback systems.} FeasibleCap is the only system that requires no head-worn display, no robot during collection, and no learned dynamics model.}}
    \vspace{-5pt}
    \centering
    \resizebox{\columnwidth}{!}{%
    \begin{tabular}{lccccc}
    \toprule
     & \rotatebox{60}{No HMD} & \rotatebox{60}{No robot} & \rotatebox{60}{Model-free} & \rotatebox{60}{Feedback} & \rotatebox{60}{Paradigm} \\
    \midrule
    ARCap~\cite{chen2024arcap} & \texttimes & \checkmark & \checkmark & V+H & Hand track. \\
    ARMADA~\cite{armada2024} & \texttimes & \checkmark & \checkmark & V & Hand track. \\
    ARMimic~\cite{armimic2025} & \texttimes & \checkmark & \checkmark & V & Hand track. \\
    FABCO~\cite{fabco2025} & \checkmark & \checkmark & \texttimes & V+H & Gripper \\
    JoyLo~\cite{joylo2025} & \checkmark & \texttimes & \checkmark & H & Teleop. \\
    \textbf{FeasibleCap} & \checkmark & \checkmark & \checkmark & V+H & Gripper \\
    \bottomrule
    \end{tabular}}
    \label{table:comparison}
    \vspace{-10pt}
\end{table}

\subsection{Inference-Time Embodiment Adaptation}

A complementary line of work addresses the embodiment gap at deployment rather than collection time. UMI-on-Air~\cite{gupta2025umionair} introduces the Embodiment-Aware Diffusion Policy, injecting MPC tracking-cost gradients into the diffusion denoising process at each step to steer generated trajectories toward the target robot's dynamic feasibility region. DPCC~\cite{romer2024dpcc} embeds model-based projections directly into the reverse diffusion loop with constraint tightening to handle model error. DDAT~\cite{ddat2025} enforces dynamically admissible trajectories via polytopic under-approximations of the reachable set at each denoising step. These inference-stage methods are complementary to FeasibleCap: they correct residual feasibility gaps at deployment but cannot prevent infeasible demonstrations from entering the training set in the first place. FeasibleCap intervenes earlier in the pipeline, improving data quality before any policy is trained.

%%%%%%%%%%%%%%%%%%%%%%%%%%%%%%%%%%%%%%%%%%%%%%%%%%%%%%%%%%%%%%%%%%%%%%%%%%%%%%%%
\section{Method}

\subsection{Problem Formulation}
\label{sec:formulation}

\begin{figure}[t]
    \centering
    \includegraphics[width=\linewidth]{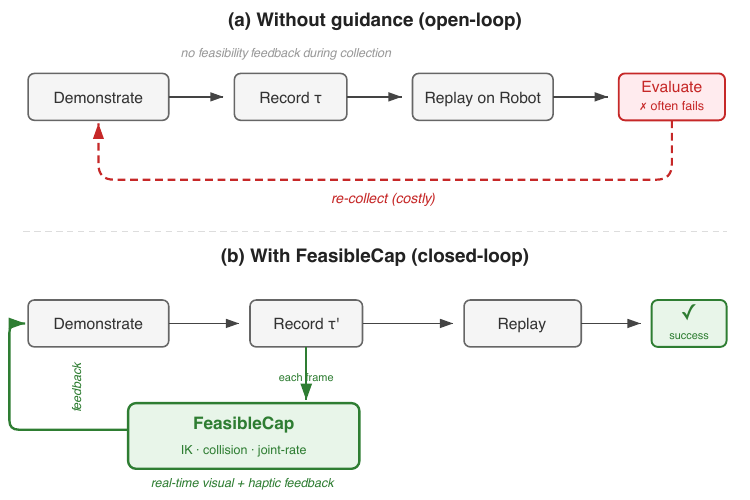}
    \caption{\footnotesize{\textbf{Open-loop vs.\ closed-loop demonstration collection.} \textit{Top}: without guidance, infeasibility is discovered only at replay time, requiring costly re-collection. \textit{Bottom}: FeasibleCap evaluates embodiment constraints in real time and feeds back visual and haptic cues, enabling the demonstrator to correct motions on the fly.}}
    \label{fig:closedloop}
    \vspace{-10pt}
\end{figure}

In the gripper-in-hand demonstration paradigm, a human demonstrator manipulates a handheld gripper to perform tasks while pose and image data are recorded for downstream policy learning. Because no robot hardware is present during collection, demonstrators receive no feedback about whether their motions lie within the target robot's executable region. Infeasibility---workspace violations, joint-rate exceedances, or collisions---is discovered only after a costly replay-and-validate loop (Fig.~\ref{fig:closedloop}, top). This open-loop workflow wastes collection effort, underrepresents challenging boundary-case motions (e.g., tossing), and provides no learning signal for demonstrators to improve their strategies.

FeasibleCap closes this loop by providing real-time embodiment constraint feedback during collection (Fig.~\ref{fig:closedloop}, bottom). Formally, let $\bm{p}_t \in SE(3)$ denote the end-effector pose produced by the demonstrator at time~$t$ and $\mathcal{M}$ the kinematic model of the target robot (loaded from a URDF). We define a pose~$\bm{p}_t$ as \textit{feasible} if and only if it simultaneously satisfies three conditions:
\begin{enumerate}
    \item \textbf{Reachability}: an inverse kinematics solution $\bm{q}_t = \text{IK}(\bm{p}_t; \mathcal{M})$ exists;
    \item \textbf{Joint-rate admissibility}: $\max_i |\dot{q}_{t,i}| / \dot{q}_{i}^{\max} \leq 1$, where $\dot{q}_{t,i}$ is the $i$-th joint velocity estimated from consecutive IK solutions;
    \item \textbf{Collision-free}: the robot configuration~$\bm{q}_t$ induces no self-collision.
\end{enumerate}
At each frame, FeasibleCap evaluates these conditions and delivers graded visual and haptic feedback to the demonstrator. The demonstrator's subsequent motion~$\bm{p}_{t+1}$ is influenced by this feedback, forming a closed-loop human-in-the-loop guidance system. The resulting trajectory $\tau = \{\bm{p}_0, \ldots, \bm{p}_T\}$ therefore contains a higher proportion of feasible frames than one collected without guidance. Crucially, FeasibleCap does not modify the recorded data: the raw pose and image streams are preserved faithfully, and the per-frame feasibility state is stored as metadata for optional downstream use (e.g., filtering or feasibility-weighted training).

% Placeholder figure: closed-loop vs open-loop

\subsection{System Overview}
\label{sec:system}

\begin{figure*}[t]
\centering
\includegraphics[width=\textwidth]{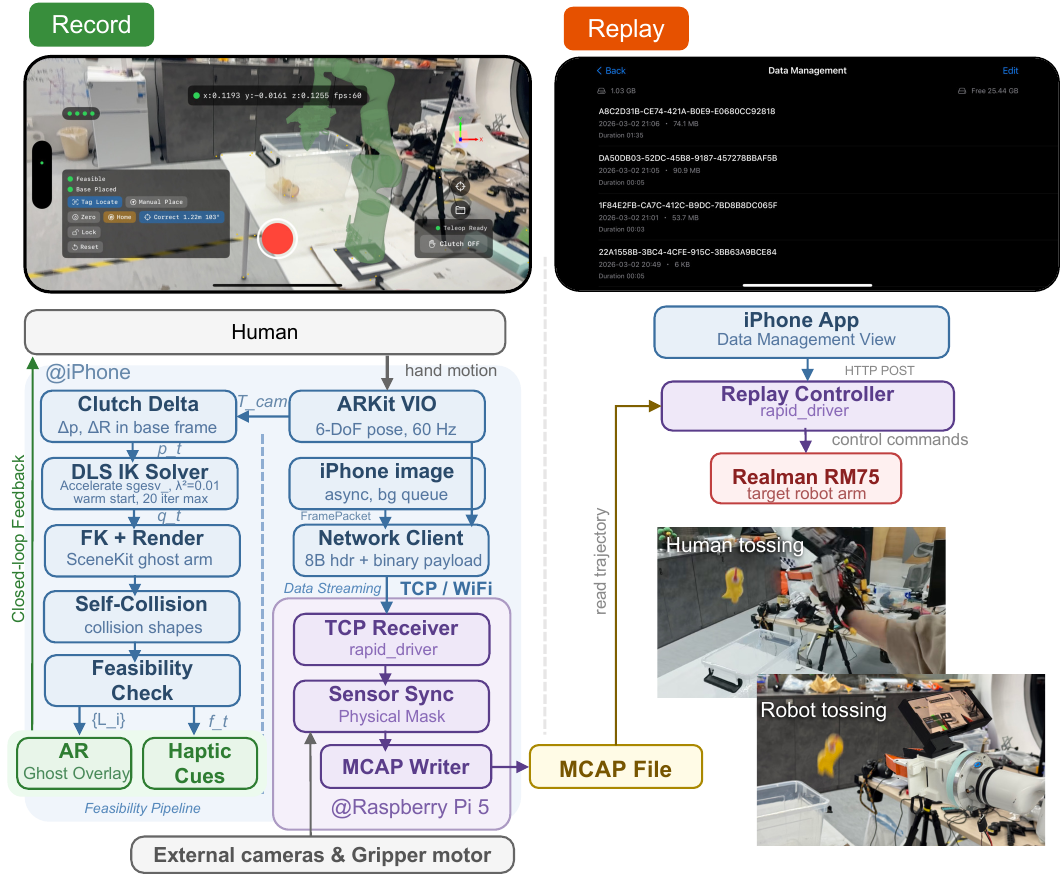}
\caption{\textbf{FeasibleCap system architecture.} \textit{Top (Record)}: the iPhone processes each ARKit frame along two parallel paths---Path~A streams compressed images and poses to the Raspberry~Pi for synchronized multi-sensor recording in MCAP format; Path~B runs the on-device feasibility pipeline (IK\,$\to$\,FK\,$\to$\,self-collision\,$\to$\,feasibility check) and closes the feedback loop via AR ghost rendering and haptic vibration. The green arrow denotes the real-time closed-loop guidance, the core contribution of this work. \textit{Bottom (Replay)}: the iPhone triggers playback; the Raspberry~Pi reads the MCAP trajectory and sends control commands to the target robot arm.}
\label{fig:system}
\vspace{-6pt}
\end{figure*}

FeasibleCap comprises three layers (Fig.~\ref{fig:system}). (1)~\textbf{Handheld device}: we build upon RAPID~\cite{yin2026rapid}, a modular handheld collection platform with built-in motor-driven gripper actuation, and mount an iPhone on its body via a 3D-printed bracket, with the camera facing outward and the screen facing the demonstrator. (2)~\textbf{iPhone application}: a native Swift application serving as the central compute and interaction hub, responsible for 6-DoF pose estimation (ARKit VIO at 60\,Hz), virtual robot IK solving and self-collision detection, AR ghost rendering and feasibility feedback, as well as recording control and data management. (3)~\textbf{Edge compute node}: a Raspberry~Pi~5 with sensor drivers written in Rust serves as the sensor synchronization and hardware coordination layer---it receives pose and image streams from the iPhone over WiFi (TCP, auto-discovered via Bonjour/mDNS), synchronizes all sensor channels (iPhone data, optional external cameras, gripper motor state via RAPID's Physical Mask mechanism), records synchronized data into MCAP files, and exposes an HTTP REST API through which the iPhone can trigger replay. During replay, the Raspberry~Pi reads recorded trajectories from the MCAP file and sends real-time control commands to the target robot arm through the manufacturer's API. The robot arm participates only in the replay stage.

\subsection{Real-Time Feasibility Guidance}
\label{sec:guidance}

The core contribution of FeasibleCap is a per-frame feasibility evaluation pipeline that runs entirely on the iPhone at 60\,Hz, enabling immediate visual and haptic feedback without any external compute.

\paragraph{Camera-to-TCP calibration.}
Because the iPhone is rigidly attached to the gripper, a fixed transform $\bm{T}_{\text{cam} \to \text{tcp}}$ relates the ARKit camera frame to the gripper's tool center point (TCP). FeasibleCap calibrates this offset through a one-shot visual alignment procedure: with the clutch disengaged (see below), the user observes both the real gripper tip and the AR ghost end-effector on screen, aligns them by hand, and presses a calibration button. This records the current relative transform as $\bm{T}_{\text{cam} \to \text{tcp}}$ for the session. Users can re-calibrate at any time if the alignment quality degrades.

\paragraph{Clutch mechanism.}
A software clutch couples or decouples the iPhone's motion from the virtual end-effector. When engaged, the iPhone pose directly drives the ghost end-effector---every hand motion is mirrored by the virtual arm. When disengaged, the ghost freezes at its last pose, allowing the user to reposition the device or inspect the ghost from different angles without generating unwanted motion. This enables users to verify the ghost's pose quality before starting a recording session.

\paragraph{Virtual robot base placement.}
Before recording begins, the user taps a point in the AR scene (via the iPhone camera view) to anchor the virtual robot's base position. Combined with the ARKit world coordinate system, this establishes the spatial relationship between the demonstrator's workspace and the target robot's kinematic frame.

\paragraph{Per-frame feasibility pipeline.}
Algorithm~\ref{alg:pipeline} summarizes the pipeline that runs entirely on the iPhone at 60\,Hz. The DLS IK solver is warm-started from the previous frame's solution to maintain convergence within the per-frame budget; it handles both 6-DoF and 7-DoF arms, with DLS naturally returning the minimum-norm solution for redundant configurations. Joint-rate ratios are smoothed over a 5-frame sliding window; the margin threshold $\tau_r{=}0.8$ triggers the \textsc{warning} state before hard-limit violation. The Yoshikawa manipulability index~\cite{yoshikawa1985manipulability} $w_t$ additionally flags near-singular configurations. The pipeline outputs three feedback states---\textsc{feasible} (green ghost, no haptic), \textsc{warning} (yellow, intermittent haptic), and \textsc{infeasible} (red, continuous haptic)---with transitions debounced over 2--3 frames to suppress flickering. SceneKit renders the ghost arm overlaid on the live camera view; self-collision is checked on non-adjacent link pairs using simplified shapes (capsules and spheres) with a 2\,cm safety margin. Haptic cues are delivered via CoreHaptics.

\begin{algorithm}[t]
\caption{Per-Frame Feasibility Pipeline}\label{alg:pipeline}
\KwIn{Robot model $\mathcal{M}$ (URDF), offset $\bm{T}_{\text{cam} \to \text{tcp}}$, thresholds $\tau_r, \tau_w$}
\For{\textnormal{each frame} $t$}{
    $\bm{T}_t^{\text{cam}} \leftarrow$ ARKit camera pose\;
    $\bm{p}_t \leftarrow \bm{T}_t^{\text{cam}} \cdot \bm{T}_{\text{cam} \to \text{tcp}}$ \tcp*[r]{target EE pose}
    $\bm{q}_t,\; e_t \leftarrow$ \texttt{dls\_ik}$(\mathcal{M},\; \bm{p}_t,\; \bm{q}_{t-1})$ \tcp*[r]{warm-started IK}
    $\{L_i\} \leftarrow$ \texttt{fk}$(\mathcal{M},\; \bm{q}_t)$\;
    Render translucent ghost arm at $\{L_i\}$\;
    $c_t \leftarrow$ \texttt{self\_collision}$(\{L_i\})$\;
    $r_t \leftarrow \max_i |\dot{q}_{t,i}| \,/\, \dot{q}_i^{\max}$ \tcp*[r]{rate ratio (5-frame window)}
    $w_t \leftarrow \sqrt{\det(\bm{J}\bm{J}^\top)}$ \tcp*[r]{manipulability}
    \uIf{$(e_t \geq \epsilon)$ \textbf{or} $c_t$ \textbf{or} $(r_t > 1)$}{
        \textsc{infeasible}: red ghost, continuous haptic\;
    }
    \uElseIf{$(r_t > \tau_r)$ \textbf{or} $(w_t < \tau_w)$}{
        \textsc{warning}: yellow ghost, intermittent haptic\;
    }
    \Else{
        \textsc{feasible}: green ghost, no haptic\;
    }
    Log $(s_t,\; e_t,\; r_t,\; c_t,\; w_t,\; \bm{p}_t,\; \textrm{image}_t)$\;
}
\end{algorithm}

\paragraph{Latency and throughput.}
On an iPhone~15~Pro~Max (Apple A17~Pro), the feasibility pipeline completes well within the 16.7\,ms budget imposed by 60\,Hz operation. Profiling over 2\,880 consecutive frames shows that the IK solver accounts for the bulk of computation at a typical latency of ${\sim}0.12$\,ms per frame, with occasional spikes up to ${\sim}2$\,ms when the DLS solver falls back to linearization; pose extraction, forward kinematics, self-collision checking, and SceneKit ghost rendering together add ${\sim}0.1$\,ms. The mean end-to-end per-frame cost is ${\sim}0.3$\,ms (worst case ${\sim}5.8$\,ms), and zero frames are dropped over the entire session. Data streaming from the iPhone to the Raspberry~Pi over WiFi incurs a round-trip latency of $5.7 \pm 1.1$\,ms. Because the feasibility pipeline runs entirely on-device and the network path is used only for asynchronous data logging, the feedback loop is decoupled from network jitter and the system maintains a stable 60\,Hz update rate throughout collection.

\subsection{Data Collection and Replay}
\label{sec:pipeline}

\paragraph{Pre-collection setup.}
The user powers on the system and waits for the Raspberry~Pi and all sensors to come online. The iPhone app verifies connectivity (all status indicators turn green), after which the user mounts the iPhone on the gripper bracket. Within the app, the user (1)~places the virtual robot base via an AR tap, (2)~optionally calibrates $\bm{T}_{\text{cam} \to \text{tcp}}$ using the visual alignment procedure, and (3)~verifies tracking quality by engaging the clutch and observing ghost responsiveness.

\paragraph{Recording.}
Once setup is complete, the app displays a record button. The user presses it to begin capture and performs the desired task while receiving real-time feasibility feedback. Pressing the button again stops recording. During recording, the iPhone streams each frame as a binary packet---containing a JPEG-compressed image, a $4\times4$ pose matrix (column-major), an ARKit timestamp, and a wall-clock timestamp---to the Raspberry~Pi over a persistent TCP connection. The Raspberry~Pi synchronizes the iPhone stream with all other connected sensors (e.g., external cameras, gripper motor encoder) through RAPID's driver layer, which supports real-time hot-plugging via the Physical Mask mechanism~\cite{yin2026rapid}, and writes all channels into an MCAP file with three primary topics: \texttt{/iphone\_pose}, \texttt{/iphone\_image}, and \texttt{/hardware\_mask}.

\paragraph{Replay.}
After collection, the user swipes to a data management view within the iPhone app. Selecting a recorded episode and pressing replay sends an HTTP request to the Raspberry~Pi's REST API. The Raspberry~Pi reads the MCAP trajectory and converts each recorded pose to a robot command: poses are expressed relative to the first frame of the trajectory and then anchored to the robot's current TCP position at replay start, with a coordinate remap between the ARKit and robot base frames. Commands are issued to the target robot arm (Realman RM75 in our experiments) at 100\,Hz, with safety velocity limits (0.25\,m/s translation, 0.5\,rad/s rotation) enforced throughout. The iPhone app displays replay progress for monitoring.
%%%%%%%%%%%%%%%%%%%%%%%%%%%%%%%%%%%%%%%%%%%%%%%%%%%%%%%%%%%%%%%%%%%%%%%%%%%%%%%%

\section{Experiments}

    We evaluate FeasibleCap on two questions: (1)~Does real-time feasibility guidance improve the quality of collected demonstrations, as measured by replay success rate? (2)~Does enforcing embodiment constraints during collection reduce cross-embodiment transferability?

    \subsection{Experimental Setup}
    \label{sec:exp_setup}

        \paragraph{Hardware.}
        All experiments use a FeasibleCap device built on the RAPID platform~\cite{yin2026rapid} with an iPhone~15~Pro~Max mounted via a 3D-printed bracket. The target robot is a Realman RM75 7-DoF arm. A Raspberry~Pi~5 handles sensor synchronization, MCAP recording, and replay command dispatch.

        \paragraph{Tasks.}
        We evaluate on two manipulation tasks:
        \begin{itemize}
            \item \textbf{Pick-and-place}: grasp a block from the table and place it into a bin. This task involves moderate workspace usage and tests basic reachability guidance.
            \item \textbf{Tossing}: grasp a block and toss it into a bin placed at a distance. This task requires fast arm motions that frequently trigger joint-rate violations, making it a stress test for FeasibleCap's velocity guidance.
        \end{itemize}

        \paragraph{Conditions.}
        We compare two conditions using identical hardware:
        \begin{itemize}
            \item \textbf{FeasibleCap} (guidance on): full AR ghost visualization with feasibility feedback (red ghost + haptic vibration on constraint violation).
            \item \textbf{Baseline} (guidance off): the same device with all feasibility feedback disabled---no ghost rendering, no haptic warnings. The device functions as a standard gripper-in-hand collection interface.
        \end{itemize}

        \paragraph{Protocol.}
        For each task and each condition, 10 demonstrations are collected and replayed on the Realman RM75.

        \paragraph{Metrics.}
        We report:
        \begin{itemize}
            \item \textbf{Replay success rate}: the fraction of demonstrations that, when replayed on the Realman RM75, successfully complete the task (block placed in / tossed into the bin). Each demonstration is replayed once. This is the primary metric.
            \item \textbf{Infeasible frame ratio}: the proportion of frames in each trajectory that violate at least one feasibility condition (Sec.~\ref{sec:formulation}). This metric is computed from the per-frame metadata logged during collection.
        \end{itemize}

    \subsection{Replay Success Rate}
    \label{sec:replay_results}

        Table~\ref{table:results} reports replay success rates across both tasks.

        % Placeholder table: replay success rates
        \begin{table}[h]
            \small
            \begin{minipage}{\linewidth}
            \caption{\footnotesize{\textbf{Replay success rates.} Each cell reports the number of demonstrations (out of 10) successfully replayed on the Realman RM75.}}
            \vspace{-5pt}
            \centering
                    \resizebox{\columnwidth}{!}{%
                    \begin{tabular}{lccc}
                    \toprule
                    Condition & Pick-and-Place & Tossing & Overall \\
                    \midrule
                    Baseline (no guidance) & 8/10 & 2/10 & 10/20 \\
                    FeasibleCap (ours) & \textbf{10/10} & \textbf{6/10} & \textbf{16/20} \\
                    \bottomrule
                    \label{table:results}
                    \end{tabular}}
            \end{minipage}
            \vspace{-10pt}
        \end{table}

        FeasibleCap achieves 10/10 replay success on pick-and-place (vs.\ 8/10 baseline) and 6/10 on tossing (vs.\ 2/10 baseline), yielding an overall rate of 16/20 compared to 10/20 without guidance. Pick-and-place is already largely feasible without guidance due to moderate speeds and workspace usage, so the headroom for improvement is small. The gain is most pronounced on tossing, where fast arm motions frequently exceed joint-rate limits: the baseline succeeds on only 2 out of 10 demonstrations, while FeasibleCap triples this to 6/10 by alerting the demonstrator to slow down or adjust the trajectory in real time. This confirms that real-time feasibility feedback is most valuable for dynamically demanding tasks where constraint violations are otherwise invisible to the demonstrator.

    \subsection{Feasibility Analysis}
    \label{sec:feasibility_analysis}

\begin{figure}[t]
            \centering
            \includegraphics[width=\linewidth]{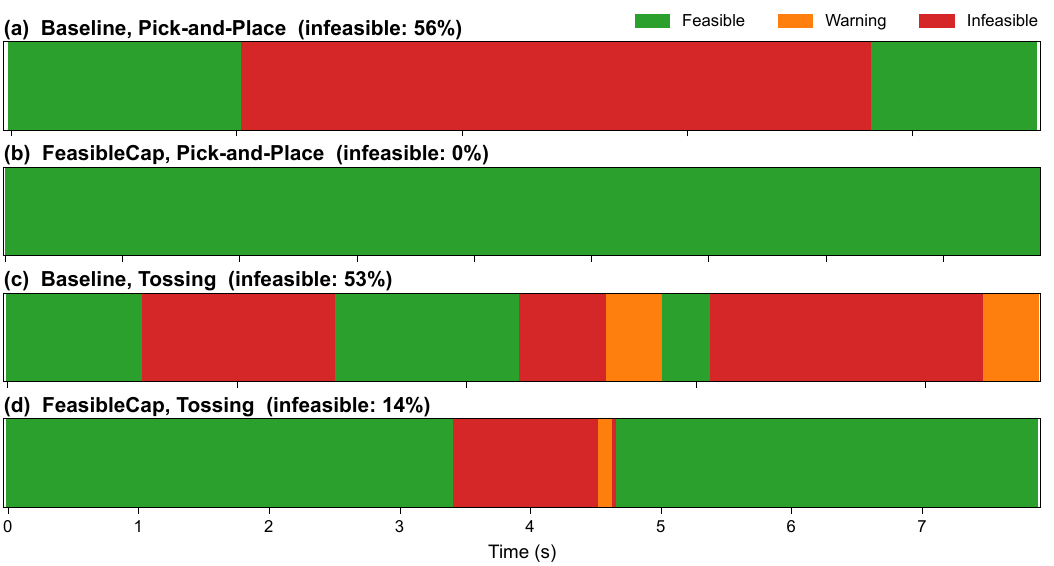}
            \caption{\footnotesize{\textbf{Per-frame feasibility timelines for representative trials.} Each bar represents one trajectory; frames are colored green (feasible), yellow (warning), or red (infeasible). (a)~Baseline, pick-and-place: 56\% infeasible. (b)~FeasibleCap, pick-and-place: 0\% infeasible. (c)~Baseline, tossing: 53\% infeasible. (d)~FeasibleCap, tossing: 14\% infeasible.}}
            \label{fig:feasibility_timeline}
            \vspace{-10pt}
        \end{figure}

        To understand how guidance affects demonstration quality at the frame level, we analyze the per-frame feasibility metadata logged during collection. Fig.~\ref{fig:feasibility_timeline} shows representative timelines from each condition, where each frame is colored green (feasible), yellow (warning), or red (infeasible).

        \paragraph{Pick-and-place.} Without guidance, baseline trajectories exhibit a mean infeasible frame ratio of $83.1 \pm 16.9\%$ across all collected trials (range: 56--100\%), indicating that the vast majority of frames violate at least one kinematic constraint. With FeasibleCap, this drops to $14.1 \pm 13.3\%$ across all guided trials---a reduction of 69~percentage points. Three FeasibleCap trials achieve 0\% infeasible frames, demonstrating that demonstrators can learn to stay entirely within the target robot's executable region when given real-time feedback. As shown in Fig.~\ref{fig:feasibility_timeline}~(a--b), the baseline trajectory is dominated by red segments, while the FeasibleCap trajectory remains fully green.

        \paragraph{Tossing.} Tossing demands fast arm motions that push joint-rate limits, making it inherently harder to keep feasible. Across all FeasibleCap tossing trials, the mean infeasible ratio is $28.7 \pm 15.6\%$ (range: 14--53\%). The best trial (14\% infeasible) shows that effective use of the guidance signal allows demonstrators to maintain mostly feasible trajectories even during high-speed motions. The worst trial (53\% infeasible) serves as a reference for what tossing looks like when the demonstrator does not fully adapt to the feedback, approaching performance comparable to unguided collection. Fig.~\ref{fig:feasibility_timeline}~(c--d) contrasts these two extremes: the 53\% trial contains frequent red segments interspersed with green, while the 14\% trial is predominantly green with only brief infeasible spikes during the toss release.

        \paragraph{Failure mode analysis.}
        Although FeasibleCap triples the tossing replay success rate from 2/10 to 6/10, four FeasibleCap tossing demonstrations still fail. Examining the per-frame feasibility logs reveals that FeasibleCap's residual failures concentrate at a single physical instant: the toss release requires a rapid wrist flick to impart projectile velocity, producing a transient joint-velocity spike that exceeds the robot's rate limits within one or two frames---too brief for the demonstrator to react to the haptic warning and correct. Fig.~\ref{fig:feasibility_timeline}(d) illustrates this pattern: the trajectory is predominantly green, but a short red band appears at mid-trajectory coinciding with the release. This spike is a physical consequence of the tossing motion itself and persists even when the demonstrator is fully attentive to the feedback.

        In contrast, baseline failures are distributed across entire trajectories and arise from a qualitatively different mechanism. Without feasibility guidance, demonstrators tend to initiate tossing with exaggerated arm swings that drive the end-effector through configurations far from the previous IK solution. The warm-started DLS solver can then converge to a distant local minimum, producing a sudden joint-configuration jump that triggers a spurious rate-limit violation. Fig.~\ref{fig:feasibility_timeline}(c) shows this pattern: infeasible frames cluster at the trajectory onset where the initial swing is most aggressive, in addition to appearing at release. With FeasibleCap active, this IK convergence failure mode largely disappears because the feedback discourages exaggerated starts.

        Contrasting the two conditions reveals a qualitative shift: without guidance, infeasible frames are scattered throughout entire trajectories (Fig.~\ref{fig:feasibility_timeline}a,\,c); with guidance, residual infeasibility is compressed into brief, physically unavoidable transients at the release instant. FeasibleCap thus converts diffuse infeasibility into concentrated infeasibility at moments where the task physics inherently conflict with the robot's rate limits---a regime that may benefit from complementary inference-time corrections~\cite{gupta2025umionair,romer2024dpcc} rather than collection-time feedback alone.

    \subsection{Cross-Embodiment Transferability}
    \label{sec:cross_embodiment}

        A natural concern is that constraining demonstrations to one robot's kinematic model may reduce the transferability of collected data to other embodiments. We test this in two directions. First, we collect demonstrations with the Franka Panda URDF as the feasibility constraint and replay them on our physical Realman RM75 (\textit{cross $\to$ real}). Second, we collect with the RM75 URDF and replay the same trajectories on Franka Panda in ManiSkill3~\cite{tao2025maniskill3} simulation (\textit{same $\to$ cross-sim}). Table~\ref{table:cross_embodiment} reports replay success rates.

        \begin{table}[h]
            \small
            \begin{minipage}{\linewidth}
            \caption{\footnotesize{\textbf{Cross-embodiment replay success rates (FeasibleCap).} ``Constraint URDF'' is the robot model used for feasibility guidance during collection. ``Replay'' indicates the robot (and environment) on which the trajectory is executed. All demonstrations are collected with FeasibleCap guidance enabled. $^\dagger$Simulation replay.}}
            \vspace{-5pt}
            \centering
                    \resizebox{\columnwidth}{!}{%
                    \begin{tabular}{llc}
                    \toprule
                    Constraint URDF & Replay & Success Rate \\
                    \midrule
                    RM75 & RM75 (same, real) & 8/10 \\
                    Franka & RM75 (cross, real) & 7/10 \\
                    \midrule
                    RM75 & Franka$^\dagger$ (cross, sim) & 8/10 \\
                    \bottomrule
                    \label{table:cross_embodiment}
                    \end{tabular}}
            \end{minipage}
            \vspace{-10pt}
        \end{table}

        When the constraint URDF differs from the replay robot (cross $\to$ real), replay success degrades only slightly: 7/10 for the Franka constraint versus 8/10 when the constraint matches the replay robot (RM75). In the reverse direction (same $\to$ cross-sim), RM75-constrained demonstrations replay on Franka in simulation at 8/10, comparable to the same-embodiment real-robot rate. Both the RM75 and Franka Panda are 7-DoF arms; these results indicate that feasibility guidance does not over-specialize demonstrations to the constraint robot, and the workspace overlap between common 7-DoF arms is large enough that trajectories feasible for one arm remain largely feasible for others.

    \subsection{Policy Training}
    \label{sec:policy_training}

        The replay success rate improvement reported in Sec.~\ref{sec:replay_results} already provides direct evidence that FeasibleCap-guided demonstrations are of higher quality: more demonstrations survive the physical replay filter, meaning the resulting dataset contains a larger proportion of executable, task-completing trajectories available for downstream training. This is the most immediate and hardware-grounded measure of data quality, as each replayed trajectory corresponds to a real execution on the target robot.

        A full end-to-end policy training comparison (e.g., training Diffusion Policy~\cite{chi2025diffusion} on guided vs.\ unguided datasets and evaluating closed-loop task success) is an important next step but falls outside the scope of this work for a practical reason: the current iPhone hardware imposes a mutual exclusion between ARKit visual-inertial odometry and wide-angle camera streaming, preventing simultaneous high-quality feasibility tracking and observation image capture through the same device. Resolving this constraint---for example by offloading observation capture to an external camera or by leveraging future iOS APIs that relax the sensor exclusion---will enable controlled policy training comparisons and is a priority for future work.

    \subsection{Limitations and Future Work}
    \label{sec:limitations}

        We note several directions for improvement.

        \textit{Feedback granularity.} The current three-state feedback already yields significant replay-rate improvements, yet it could be extended to a continuous feasibility score combining manipulability, joint-rate margins, and collision clearance, enabling proportional visual and haptic cues that help demonstrators optimize trajectories rather than merely avoid violations.

        \textit{Tracking robustness.} ARKit's visual-inertial odometry relies on distinctive visual features for accurate pose estimation. Task scenes must therefore contain sufficient texture and geometric detail within the camera's field of view; in feature-sparse environments the tracker can degrade, causing drift in the ghost-arm overlay. Selecting viewpoints that keep rich background features visible during collection is important for maintaining tracking quality.

        \textit{Learning cost.} Operating with real-time feasibility feedback introduces a learning curve: demonstrators must attend to visual and haptic cues while performing the task, which can initially slow collection speed compared to unconstrained recording. In our experiments operators adapted within a few trials, but the trade-off between demonstration quality and collection throughput remains a practical consideration.

        \textit{Broader evaluation.} Our experiments span multiple tasks and robot platforms, validating the system's generality; a larger-scale user study and end-to-end policy training comparisons would further strengthen the conclusions. Extending the system to bimanual manipulation is another natural next step.

%%%%%%%%%%%%%%%%%%%%%%%%%%%%%%%%%%%%%%%%%%%%%%%%%%%%%%%%%%%%%%%%%%%%%%%%%%%%%%%%

\section{Conclusion}

    We presented FeasibleCap, a gripper-in-hand data collection system that brings real-time embodiment constraint guidance into robot-free demonstration capture. By evaluating reachability, joint-rate limits, and self-collisions on-device at 60\,Hz and delivering immediate visual and haptic feedback, FeasibleCap enables demonstrators to correct infeasible motions during collection rather than discovering failures at replay time. Experiments on pick-and-place and tossing tasks show that guidance improves replay success rates, with the largest gains on tossing where joint-rate constraints are most sensitive. Per-frame feasibility analysis confirms that replay failures correlate strongly with elevated infeasible frame ratios, validating the causal mechanism behind the improvement. Cross-embodiment experiments further indicate that constraining demonstrations to one robot's kinematic model does not sacrifice transferability to other platforms. Future work includes extending the feedback from discrete to continuous feasibility scores and supporting bimanual collection.

%%%%%%%%%%%%%%%%%%%%%%%%%%%%%%%%%%%%%%%%%%%%%%%%%%%%%%%%%%%%%%%%%%%%%%%%%%%%%%%%


%%%%%%%%%%%%%%%%%%%%%%%%%%%%%%%%%%%%%%%%%%%%%%%%%%%%%%%%%%%%%%%%%%%%%%%%%%%%%%%%

%\section*{ACKNOWLEDGMENTS}
%
%    \todo{Acknowledge funding sources and collaborators.}

%%%%%%%%%%%%%%%%%%%%%%%%%%%%%%%%%%%%%%%%%%%%%%%%%%%%%%%%%%%%%%%%%%%%%%%%%%%%%%%%

\bibliographystyle{IEEEtran}
\bibliography{refs.bib}

\end{document}